\documentclass{article} %
\usepackage{iclr2025_conference,times}

\usepackage{hyperref}
\usepackage{url}
\usepackage{amsmath,amsfonts,bm}
\usepackage{booktabs}
\usepackage{graphicx} 
\usepackage{colortbl}
\usepackage{caption}
\usepackage{subcaption}
\usepackage{xspace}
\definecolor{mygray}{gray}{.9}
\makeatletter
\DeclareRobustCommand\onedot{\futurelet\@let@token\@onedot}
\def\@onedot{\ifx\@let@token.\else.\null\fi\xspace}

\def\ie{\emph{i.e}\onedot}

\makeatother

\newcommand{\app}{\raise.17ex\hbox{$\scriptstyle\sim$}}
\makeatletter\renewcommand\paragraph{\@startsection{paragraph}{4}{\z@}
  {.15em \@plus1ex \@minus.2ex}{-.5em}{\normalfont\normalsize\bfseries}}\makeatother
  
\title{M-VAR: Decoupled Scale-wise Autoregressive Modeling for High-Quality Image Generation}

\author{Sucheng Ren$^1$,~~~ Yaodong Yu$^2$,~~ Nataniel Ruiz$^3$,~~ Feng Wang$^1$,~~ Alan Yuille$^1$,~~ Cihang Xie$^4$\\
$^1$ Johns Hopkins University~~~
$^2$ University of California, Berkeley ~~~
$^3$ Google~~~
$^4$ UC Santa Cruz
}

\iclrfinalcopy %
\begin{document}

\maketitle
{%
\begin{center}
    \centering
    \captionsetup{type=figure}
    \includegraphics[width=\linewidth]{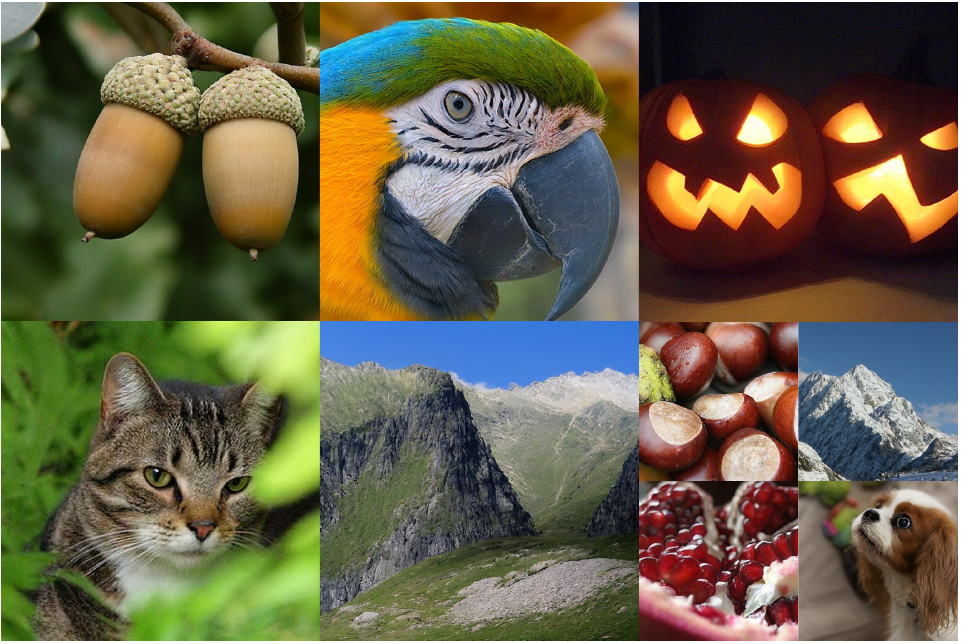}
    \vspace{-1.7em}
    \captionof{figure}{ \textbf{Generated 512$\times$512 and 256$\times$256 samples from our M-VAR trained on ImageNet.}
    }
    \label{fig:teaser1}
\end{center}%
}
\begin{abstract}

There exists recent work in computer vision, named VAR, that proposes a new autoregressive paradigm for image generation. Diverging from the vanilla next-token prediction, VAR structurally reformulates the image generation into a coarse to fine next-scale prediction. 
In this paper, we show that this scale-wise autoregressive framework can be effectively decoupled into \textit{intra-scale modeling}, which captures local spatial dependencies within each scale, and \textit{inter-scale modeling}, which models cross-scale relationships progressively from coarse-to-fine scales.
This decoupling structure allows to rebuild VAR in a more computationally efficient manner. Specifically, for intra-scale modeling --- crucial for generating high-fidelity images --- we retain the original bidirectional self-attention design to ensure comprehensive modeling; for inter-scale modeling, which semantically connects different scales but is computationally intensive, we apply linear-complexity mechanisms like Mamba to substantially reduce computational overhead. 
We term this new framework M-VAR. Extensive experiments demonstrate that our method outperforms existing models in both image quality and generation speed. For example, our 1.5B model, with fewer parameters and faster inference speed, outperforms the largest VAR-d30-2B. Moreover, our largest model M-VAR-d32 impressively registers 1.78 FID on ImageNet 256$\times$256 and outperforms the prior-art autoregressive models LlamaGen/VAR by 0.4/0.19 and popular diffusion models LDM/DiT by 1.82/0.49, respectively. Code is avaiable at \url{https://github.com/OliverRensu/MVAR}.
\end{abstract}

\section{Introduction}
Autoregressive models~\citep{gpt,gpt3} have been instrumental in advancing the field of natural language processing (NLP). By modeling the probability distribution of a token given the preceding ones, these models can generate coherent and contextually relevant text. Prominent examples like GPT-3~\citep{gpt3} and its successors~\citep{chatgpt,gpt4} have demonstrated remarkable capabilities in language understanding and generation, setting new benchmarks across various NLP applications.

\begin{figure}[t!]
    \centering
    \includegraphics[width=0.65\linewidth]{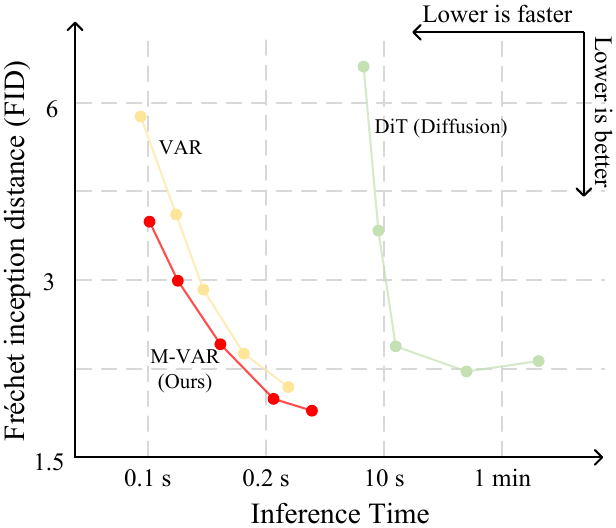}
    \caption{Fréchet inception distance (FID) on 256$\times$256 image generation. Our M-VAR-1.5B model outperforms the largest 2B VAR-d30 with fewer parameters and faster inference speed. Our largest M-VAR-3B achieves 1.78 FID.   }
    \label{fig:teaser}
\end{figure}

Building upon the success in NLP, the autoregressive modeling paradigm~\citep{parti,llamagen,pixelcnn} has also been extended to computer vision for image generation tasks, aiming to generate high-fidelity images by predicting visual content in a sequential manner. 
Recently, VAR~\citep{var} has further enhanced this image autoregressive pipeline by structurally reformulating the learning target into a coarse-to-fine ``next-scale prediction'', which innately introduces strong semantics to interconnecting tokens along scales. As demonstrated in the VAR paper, this pipeline exhibits much stronger scalability and can achieve competitive, sometimes even superior, performance compared to advanced diffusion models.

This paper aims to further optimize VAR's computation structure. Our key insight lies in decoupling VAR's cross-scale autoregressive modeling into two distinct parts: intra-scale modeling and inter-scale modeling. 
Specifically, intra-scale modeling involves bidirectionally modeling multiple tokens within each scale, capturing intricate spatial dependencies and preserving the 2D structure of images. In contrast, inter-scale modeling focuses on unidirectional causality between scales by sequentially progressing from coarse to fine resolutions ---  each finer scale is generated conditioned on all preceding coarser scales, ensuring that global structures guide the refinement of local details. 
Notably, the sequence length involved in inter-scale modeling is much longer than that of intra-scale modeling, resulting in significantly higher computational costs. But meanwhile, our analysis of attention scores for both intra-scale and inter-scale interactions (as discussed in Sec. \ref{sec:analysis}) suggests a contrasting reality: intra-scale interactions dominate the model's attention distribution, while inter-scale interactions contribute significantly less.

Motivated by the observations above, we propose to develop a more customized computation configuration for VAR. For the intra-scale component, given the much shorter sequence lengths within each scale and its significant contribution to the model's attention distribution, we retain the bidirectional attention mechanism to fully capture comprehensive spatial dependencies. This ensures that local spatial relationships and fine-grained details are effectively modeled at a reasonable computational overhead. Conversely, for the inter-scale component, which involves much longer sequences but demands relatively less comprehensiveness in modeling global relationships, we adopt Mamba~\citep{mamba1,mamba2}, a linear-complexity mechanism, to handle such inter-scale dependencies efficiently.

By segregating these two modeling modules and applying appropriate mechanisms to each, our approach significantly reduces computational complexity while preserving the model’s ability to maintain 2D spatial coherence and unidirectional coarse-to-fine consistency, making it well-suited for high-quality image generation. As shown in Figure \ref{fig:teaser}, our proposed framework, which we term M-VAR, outperforms existing models in both image quality and inference speed. For instance, comparing to the largest VAR model, which has 2B parameters and attains an FID score of 1.97, our 1.5B parameter M-VAR model achieves an FID score of 1.93 with much fewer parameters and 1.2$\times$ faster inference speed. M-VAR also scales well --- our largest model, M-VAR-d32, achieves an impressive FID score of 1.78 on ImageNet at $256\times256$ resolution, outperforming the prior best autoregressive models LlamaGen by 0.4 and VAR by 0.19, respectively, and well-known diffusion models LDM by 1.82 and DiT by 0.49, respectively.

\section{Related Work}
\subsection{Visual Generation}
Visual generation can generally be split into three categories: 1) Diffusion models~\citep{adm,ldm} which treat visual generation as the reverse process of the diffusion process; 2) Mask prediction models~\citep{maskgit} which follow BERT-style~\citep{bert} language model to generate images by predicting mask tokens; and 3) Autoregressive models which generate images by predicting the next pixel/token/scale in a sequence. We focus on the last one in this paper.

The pioneering method that brings autoregressive modeling into visual generation is PixelCNN~\citep{pixelcnn}, which models images by predicting the discrete probability distributions of raw pixel values, effectively capturing all dependencies within an image. Building on this foundation, VQGAN~\citep{vqgan} advances the field by applying autoregressive learning within the latent space of VQVAE~\citep{vqvae2}, simplifying the data representation for more efficient modeling. The RQ Transformer~\citep{rq} introduces a novel technique using a fixed-size codebook to approximate an image's feature map with stacked discrete codes, forecasting the next quantized feature vectors by predicting subsequent code stacks. Parti~\citep{parti} takes a different route by framing image generation as a sequence-to-sequence modeling task akin to machine translation, using sequences of image tokens as targets instead of text tokens, and thus capitalizing on the significant advancements made in large language models through data and model scaling. LlamaGen~\citep{llamagen} further extends this concept by applying the traditional "next-token prediction" paradigm of large language models to visual generation, demonstrating that standard autoregressive models like Llama can achieve state-of-the-art image generation performance when appropriately scaled, even without considering specific inductive biases for visual signals. Lastly, departing from the conventional raster-scan ``next-token prediction'' method, the recent work VAR~\citep{var} offers a new perspective by developing a coarse-to-fine ``next-scale prediction'' strategy for visual autoregressive modeling. Our work is a followup of VAR, aiming to making the whole framework more efficient.

\subsection{Mamba}
State-space models (SSMs)~\citep{ssm1,ssm2} have recently emerged as a compelling alternative to Transformers~\citep{vaswani2017attention}, which employ hidden states to capture long-range dependencies  efficiently. The latest advancement in this domain is Mamba~\citep{mamba1,mamba2}, a sophisticated SSM that introduces data-dependent layers with expanded hidden states. Empirically, Mamba is able to construct a versatile language model backbone that not only rivals Transformers across various scales but also maintains linear scalability with respect to sequence length.

Building on Mamba's success in NLP, its application has been quickly extended to computer vision tasks. Vision Mamba (Vim)~\citep{vim} utilizes pure Mamba layers within Vim blocks, leveraging both forward and backward scans to model bidirectional representations. This approach effectively addresses the direction-sensitive limitations of the original Mamba model. Additionally, ARM~\citep{arm} pioneers the integration of autoregressive pretraining with Mamba in the vision domain.

For image generation, Diffusion Mamba (DiM)~\citep{dim} combines the efficiency of the Mamba sequence model with diffusion processes to achieve high-resolution image synthesis. Specifically, DiM employs multi-directional scans, introduces learnable padding tokens, and enhances local features to adeptly manage two-dimensional signal processing. AiM~\citep{aim} further advances this by replacing Transformers with Mamba for autoregressive image generation, following methodologies similar to LlamaGen~\citep{llamagen}. However, these existing methods typically apply Mamba to sequences with the length merely up to 256. Our proposed M-VAR model further unleash Mamba's \emph{true} capability in capturing super-long-sequence dependencies, empirically up to 2,240 visual tokens. This \app$10\times$ increase in sequence length can further highlight the benefits brought by Mamba's efficiency in modeling long sequences within vision applications.

\section{Method}
\subsection{Preliminary: Autoregressive Modeling}
\paragraph{Autoregressive modeling in natural language processing.} Given a set of corpus $\mathcal{U} = \{u_1, ..., u_n\}$, autoregressive modeling predicts next words based on all preceding words:
\begin{equation}
    p(u) = \prod\limits_{i=1}^{n} p(u_i|u_1,...,u_{i-1}, \Theta)
\end{equation}
Autoregressive modeling minimize the negative log-likelihood of each word $u_i$ given all preceding words from $u_1$ to $u_{i-1}$:
\begin{equation}
    \mathcal{L} =  - \log~ p(u)
\end{equation}
This strategy leads to the success of a large language model~\cite{llama,chatgpt,gpt4}

\paragraph{Token-wise autoregressive modeling in computer vision.} From language to image, to apply autoregressive pertaining, image tokenization via vector-quantization transfers 2D images $X\in \mathcal{R}^{H\times W \times C}$ into 2D tokens and then flatten these tokens into 1D token sequences $X=\{x_1, x_2, ...,x_n\}$:
\begin{equation}
    \mathcal{L} =  -\sum\limits_{i=1}^{N} \log~ p(x_i|x_1,...,x_{i-1}, \Theta)
\end{equation}

However, such the flatten operation breaks the inherent 2D structure of an image. To address this issue, VAR~\citep{var} proposes to perform scale-wise autoregressive modeling to keep the 2D structure, as detailed next.

\paragraph{Scale-wise autoregressive modeling.}  Instead of tokenizing image into a sequence of tokens, VAR tokenizes the image into multi-scale token maps $S=\{s_1, ..., s_n\}$, where $s_i$ is the token map with the resolution of $h_i\times w_i$ downsampling from $s_n \in \mathcal{R}^{h_n\times w_n}$. Compared to $x_i$ which only contains one token and breaks the 2D structure, $s_i$ contains $h_i\times w_i$ tokens and is able to maintain the 2D structure. Then, the corresponding autoregressive model is reformed to:
\begin{equation}
    \mathcal{L} =  -\sum\limits_{i=1}^{N} \log~ p(s_i|s_1,...,s_{i-1}, \Theta)
\end{equation}
Note that the sequence $S$ of multiple scales is much longer than each scale ($s_1, ..., s_n$). VAR by default utilizes attention and Transformer to instantiate this modeling --- \ie, to generate the $ i_{th}$ scale, VAR needs to attend all preceding scales, ranging from the first one to the  $(i-1)_{th}$ one, and predict all the $h_i\times w_i$ tokens in parallel inside the $ i_{th}$ scale.

\subsection{Decouple Scale-wise Autoregressive  Modeling}
\label{sec:analysis}
We note the attention in VAR can actually be separated into two parts: 1) the first part bidirectionally attends the intra-scale information, where the sequence length is much shorter; 2) the second part unidirectionally attends from the coarse scale to the fine scale, where the sequence length is much longer. 
We show the statistics of these two parts of attention modeling, including the attention score and the associated computation cost, in Table \ref{tab:score}. 

Firstly, we note that intra-scale attention scores account for 79.6\% of the total attention scores in 256$\times$256 image generation and 77.1\% in 512$\times$512 image generation. This dominance of intra-scale attention suggests that capturing fine-grained details within the same scale is crucial for high-quality image synthesis. More interestingly, we observe that the associated computation cost presents a contrasting scenario --- despite intra-scale attention contributing the most to the attention scores, it only consumes 23.9\% of the computation cost for 256$\times$256 images and 30.3\% for 512$\times$512 images; In stark contrast, inter-scale attention, which accounts for a much smaller portion of the attention scores (20.4\% for 256$\times$256 and 22.9\% for 512$\times$512 images), is responsible for the majority of the computation cost, \ie, 76.1\% and 69.7\% respectively. The disparity between the attention scores and computation cost highlights an inefficiency in the current attention configuration in VAR. 

Based on this observation, we propose a novel approach to optimize the computational efficiency of the scale-wise autoregressive image generation models. Specifically, we retain the standard attention mechanisms for intra-scale interactions, 
but, importantly, \textit{employ Mamba for inter-scale interactions}. The main motivation of this change is that Mamba is designed to handle long-range interactions efficiently that scales linearly with the sequence length, as opposed to the quadratic scaling of traditional attention mechanisms. This property makes Mamba particularly suitable for modeling inter-scale relationships, where the computational cost is otherwise prohibitive. We name this new method M-VAR.

\begin{table}[t]
    \centering
    \begin{tabular}{c|c|c}
    \toprule
        Attention   Mode      & Attention Score & Computation Cost \\
        \midrule
        \multicolumn{3}{c}{\emph{256$\times$256 Image Generation}}\\
        Intra Scale & 79.6\% &  23.9\% \\
        Inter Scale & 20.4\% &   76.1\%  \\
        \midrule
        \multicolumn{3}{c}{\emph{512$\times$512 Image Generation}}\\
        Intra Scale & 77.1\% &  30.3\% \\
        Inter Scale & 22.9\% &   69.7\%  \\
    \bottomrule
    \end{tabular}
    \caption{The statistics of attention score and computation cost of the attention in VAR.}
    \label{tab:score}
\end{table}

\begin{figure}[t]
    \centering
    \includegraphics[width=0.65\linewidth]{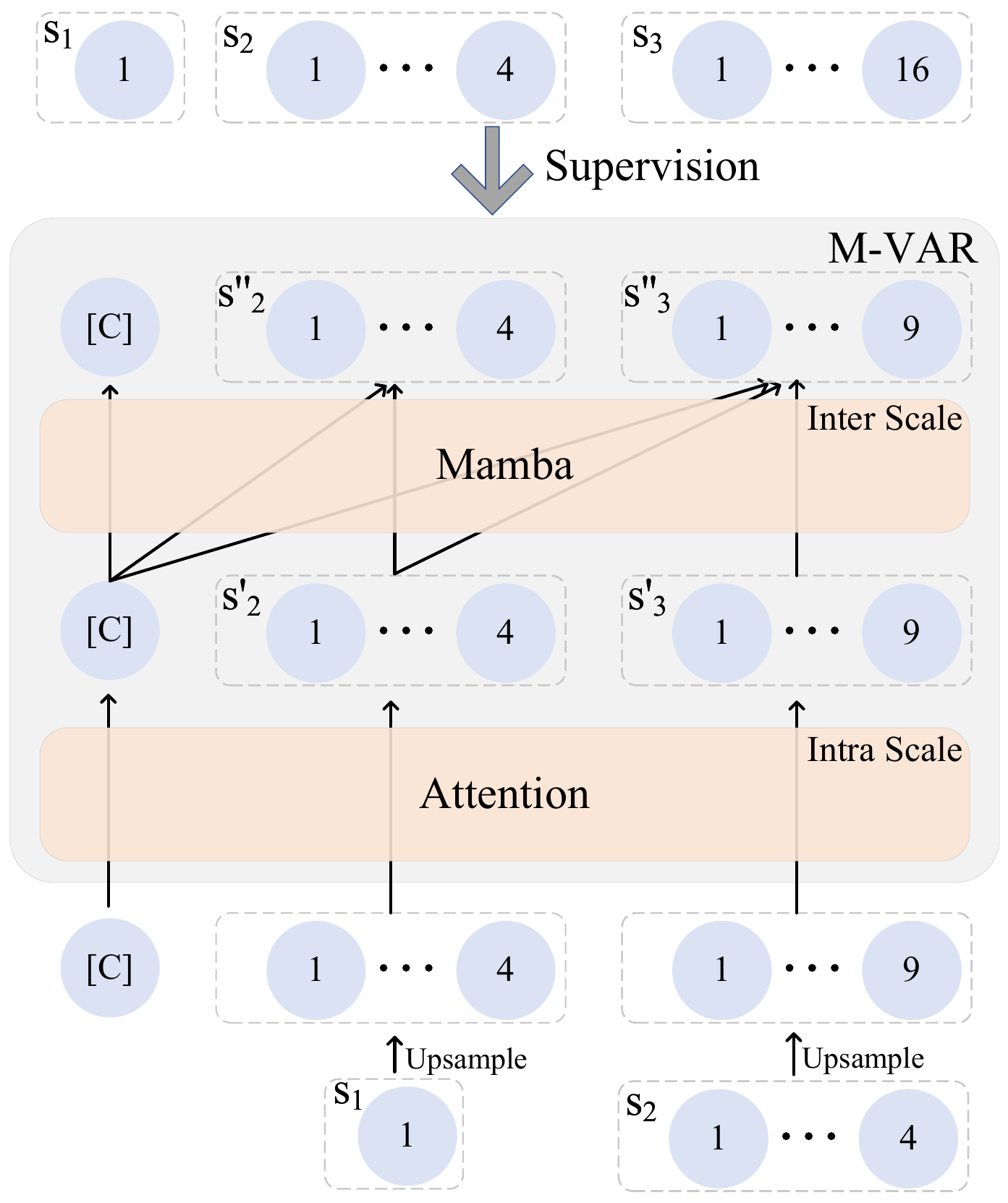}
    \caption{An overview of M-VAR. M-VAR takes the input sequence of $\{[C], s_1, ..., s_{n-1}\}$ to predict $\{s_1, ..., s_{n}\}$ where $[C]$ is the condition token. The model first divides the input into different scales and applies the standard attention mechanism to capture intra-scale spatial correlations. It then utilizes Mamba to autoregressively model inter-scale dependencies, enabling coherent and efficient multi-scale image generation.}
    \label{fig:method}
\end{figure}

Formally, as illustrated in Figure \ref{fig:method}, M-VAR models inputs in the following two steps. First, given an image with multiple scales $S=[s_1, ..., s_n]$, 
 we apply the standard attention mechanism independently to each scale for capturing the fine-grained details and local dependencies:
\begin{equation}
    S^{'}=[s^{'}_1, ...,s^{'}_n]=[Attn(C), Attn(Upsample(s_1)), ..., Attn(Upsample(s_{n-1}))]
\end{equation}
Here, $Attn$ represents the standard attention, which produces the intra-scale representation, and $C$ is the condition token. All attention share the same parameters but process each scale independently. This design choice ensures consistency across scales and reduces the overall model complexity. For efficient implementation, we adopt FlashAttention~\citep{flash1,flash2} to perform the intra-scale attention in parallel.

After obtaining the intra-scale representations $S^{'}$, modeling the relation between different scales becomes crucial for ensuring global coherence and coarse-to-fine consistency in the generated images. However, as previously discussed, traditional attention mechanisms are computationally expensive for inter-scale interactions due to their quadratic complexity, and we adopt the Mamba model with linear complexity, as
\begin{equation}
    S^{''} =[s^{''}_1, ...,s^{''}_n]= Mamba(Concat([s^{'}_1, ..., s^{'}_{n}]))
\end{equation}
By concatenating $s^{'}$ from all scales into a single sequence, Mamba efficiently processes the combined representations, capturing the essential inter-scale interactions without incurring heavy computational burden.

\section{Experiment}

\subsection{ImageNet 256×256 conditional generation}
Following the same setup in \citet{var}, we first train M-VAR on ImageNet~\citep{deng2009imagenet} for 256$\times$256 conditional generation. We design multiple model variants with depths of 12,  16, 20, 24, and 32 layers. 

We compare our M-VAR with previous state-of-the-art generative adversarial nets (GANs), diffusion models, autoregressive models, mask prediction models. As shown in Table \ref{tab:main}, Our M-VAR offers a balanced synergy of high image quality and computational efficiency. For example, M-VAR outperforms GANs in terms of image fidelity and diversity while maintaining comparable inference speeds. Compared to diffusion models, M-VAR delivers comparable or even superior image quality with significantly reduced inference time. Against token-wise autoregressive and mask prediction models, our M-VAR achieves better performance metrics with fewer steps and faster inference times.

Compared with the most related VAR, our M-VAR demonstrates significant advancements in both performance and efficiency. Across various depths, M-VAR consistently achieves lower Fréchet Inception Distance (FID) scores and higher Inception Scores (IS), indicating superior image quality and diversity. Specifically, M-VAR-d24 attains an FID of 1.93 and an IS of 320.7 with 1.5 billion parameters, which already surpasses the largest VAR model, VAR-d30, with 25\% fewer parameters and 14\% faster inference speed. M-VAR also shows compelling scalability --- our largest model, M-VAR-d32, achieves state-of-the-art performance with an FID of 1.78 and an IS of 331.2, utilizing \app3 billion parameters. These results highlight the effectiveness of our approach in integrating intra-scale attention with Mamba for inter-scale modeling, leading to superior image generation quality and computational efficiency compared to existing models. 
In additional to these quantitative numbers, we also show more qualitative results in Figure \ref{fig:qualit}.

\begin{table}[t]
\renewcommand\arraystretch{1.05}
\centering
\setlength{\tabcolsep}{2.5mm}{}
\small
{
\caption{
\textbf{Generative model comparison on class-conditional ImageNet 256$\times$256}.
Metrics include Fréchet inception distance (FID), inception score (IS), precision (Pre) and recall (rec).
Step: the number of model runs needed to generate an image.
Time: the relative inference time of M-VAR.
}\label{tab:main}

\begin{tabular}{l|cc|cc|cc|c}
\toprule
 Model          & FID$\downarrow$ & IS$\uparrow$ & Pre$\uparrow$ & Rec$\uparrow$ & Param & Step & Time \\
\midrule
\multicolumn{8}{c}{\emph{Generative Adversarial Net (GAN)}} \\
 BigGAN~\citep{biggan}  & 6.95  & 224.5       & 0.89 & 0.38 & 112M & 1    & $-$    \\
 GigaGAN~\citep{gigagan}     & 3.45  & 225.5       & 0.84 & 0.61 & 569M & 1    & $-$ \\
 StyleGan-XL~\citep{stylegan-xl}  & 2.30  & 265.1       & 0.78 & 0.53 & 166M & 1    & 0.2 \\
\midrule
\multicolumn{8}{c}{\emph{Diffusion}} \\
 ADM~\citep{adm}         & 10.94 & 101.0        & 0.69 & 0.63 & 554M & 250  & 118 \\
 CDM~\citep{cdm}         & 4.88  & 158.7       & $-$  & $-$  & $-$  & 8100 & $-$    \\
 LDM-4-G~\citep{ldm}     & 3.60  & 247.7       & $-$  & $-$  & 400M & 250  & $-$    \\
 DiT-L/2~\citep{dit}     & 5.02  & 167.2       & 0.75 & 0.57 & 458M & 250  & 2     \\
 DiT-XL/2~\citep{dit}    & 2.27  & 278.2       & 0.83 & 0.57 & 675M & 250  & 2     \\
L-DiT-3B~\citep{dit-github}    & 2.10  & 304.4       & 0.82 & 0.60 & 3.0B & 250  & $>$32     \\
L-DiT-7B~\citep{dit-github}    & 2.28  & 316.2       & 0.83 & 0.58 & 7.0B & 250  & $>$32     \\
\midrule
\multicolumn{8}{c}{\emph{Mask Prediction}} \\
 MaskGIT~\citep{maskgit}     & 6.18  & 182.1        & 0.80 & 0.51 & 227M & 8    & 0.4  \\
 RCG (cond.)~\citep{rcg}  & 3.49  & 215.5        & $-$  & $-$  & 502M & 20  & 1.4  \\
\midrule
\multicolumn{8}{c}{\emph{Token-wise Autoregressive}} \\
 VQVAE-2$^\dag$~\citep{vqvae2} & 31.11           & $\sim$45     & 0.36           & 0.57          & 13.5B    & 5120    & $-$  \\
 VQGAN$^\dag$~\citep{vqgan} & 18.65 & 80.4         & 0.78 & 0.26 & 227M & 256  & 7   \\
 VQGAN~\citep{vqgan}       & 15.78 & 74.3   & $-$  & $-$  & 1.4B & 256  & 17     \\

 ViTVQ~\citep{vit-vqgan}& 4.17  & 175.1  & $-$  & $-$  & 1.7B & 1024  & $>$17     \\

RQTran.~\citep{rq}        & 7.55  & 134.0  & $-$  & $-$  & 3.8B & 68  & 15    \\

LlamaGen-3B~\citep{llamagen}& 2.18& 263.33 &0.81& 0.58 &3.1B&576& -\\
\midrule
\multicolumn{8}{c}{\emph{Scale-wise Autoregressive}} \\
VAR-$d12$~\citep{var}       &  5.81 &201.3  & 0.81 & 0.45 & 132M  & 10   &    0.2  \\
 \rowcolor{cyan!10}
M-VAR-$d12$       & 4.19  & 234.8 &  0.83&0.48 & 198M  & 10   &  0.2   \\
VAR-$d16$~\citep{var}        & 3.55  & 280.4 & 0.84 & 0.51 & 310M & 10   & 0.2     \\
 \rowcolor{cyan!10}
M-VAR-$d16$       & 3.07  & 294.6 & 0.84&0.53  &  464M  & 10   &0.2 \\
VAR-$d20$~\citep{var}        & 2.95  & 302.6 & 0.83 & 0.56 & 600M & 10   &  0.3   \\
 \rowcolor{cyan!10}
M-VAR-$d20$       & 2.41  & 308.4 & 0.85 & 0.58 & 900M & 10   &0.4 \\
 VAR-$d24$~\citep{var}        & 2.33  & 312.9 & 0.82 & 0.59 & 1.0B & 10   & 0.5   \\
 \rowcolor{cyan!10}
 M-VAR-$d24$       & 1.93 & 320.7 & 0.83 & 0.59 & 1.5B & 10   &   0.6  \\
VAR-$d30$~\citep{var}        & 1.97  & 323.1 & 0.82 & 0.59 & 2.0B & 10   & 0.7      \\
 \rowcolor{cyan!10}
M-VAR-$d32$       & 1.78 & 331.2 & 0.83 & 0.61 & 3.0B & 10   &  1  \\

\bottomrule
\end{tabular}
}

\end{table}

\begin{table}[t]
\renewcommand\arraystretch{1.05}
\centering
\setlength{\tabcolsep}{2.5mm}{}
\small
{
\caption{
\textbf{Generative model comparison on class-conditional ImageNet 256$\times$256 with rejection sampling}.
}\label{tab:rej}

\begin{tabular}{l|c|cc}
\toprule
 Model      &Params    & FID$\downarrow$ & IS$\uparrow$  \\
\midrule
 ViTVQ~\citep{vit-vqgan}& 1.7B & 3.04  & 227.4       \\
 RQTran.~\citep{rq}     & 3.8B & 3.80  & 323.7     \\
 VQGAN~\citep{vqgan}   & 1.4B & 5.20  & 280.3      \\
 VAR-d30~\citep{var} & 2.0B&1.80& 343.2\\
  \rowcolor{cyan!10}
M-VAR-d32& 3.0B&1.67& 361.5\\
\bottomrule
\end{tabular}
}
\end{table}

To further enhance the generation results, as shown in Table \ref{tab:rej}, we compare our M-VAR-d32 with other state-of-the-art methods using rejection sampling~\cite{rej1,rej2} on class-conditional ImageNet 256$\times$256. Our M-VAR-d32 achieves an FID of 1.63 and an IS of 361.5, outperforming all compared models. Specifically, it surpasses the prior art VAR-d30 by 0.1 in FID   and by 11.3 in IS. Additionally, M-VAR-d32 demonstrates significant improvements over ViTVQ, RQTransformer, and VQGAN by an FID of 1.41, 2.17, 3.57 respectively. These results highlight the effectiveness of our approach in achieving superior image generation quality under rejection sampling, affirming the advancements of M-VAR in image generation.

\subsection{ImageNet 512×512 conditional generation}

\begin{figure}[t]
    \centering
    \includegraphics[width=\linewidth]{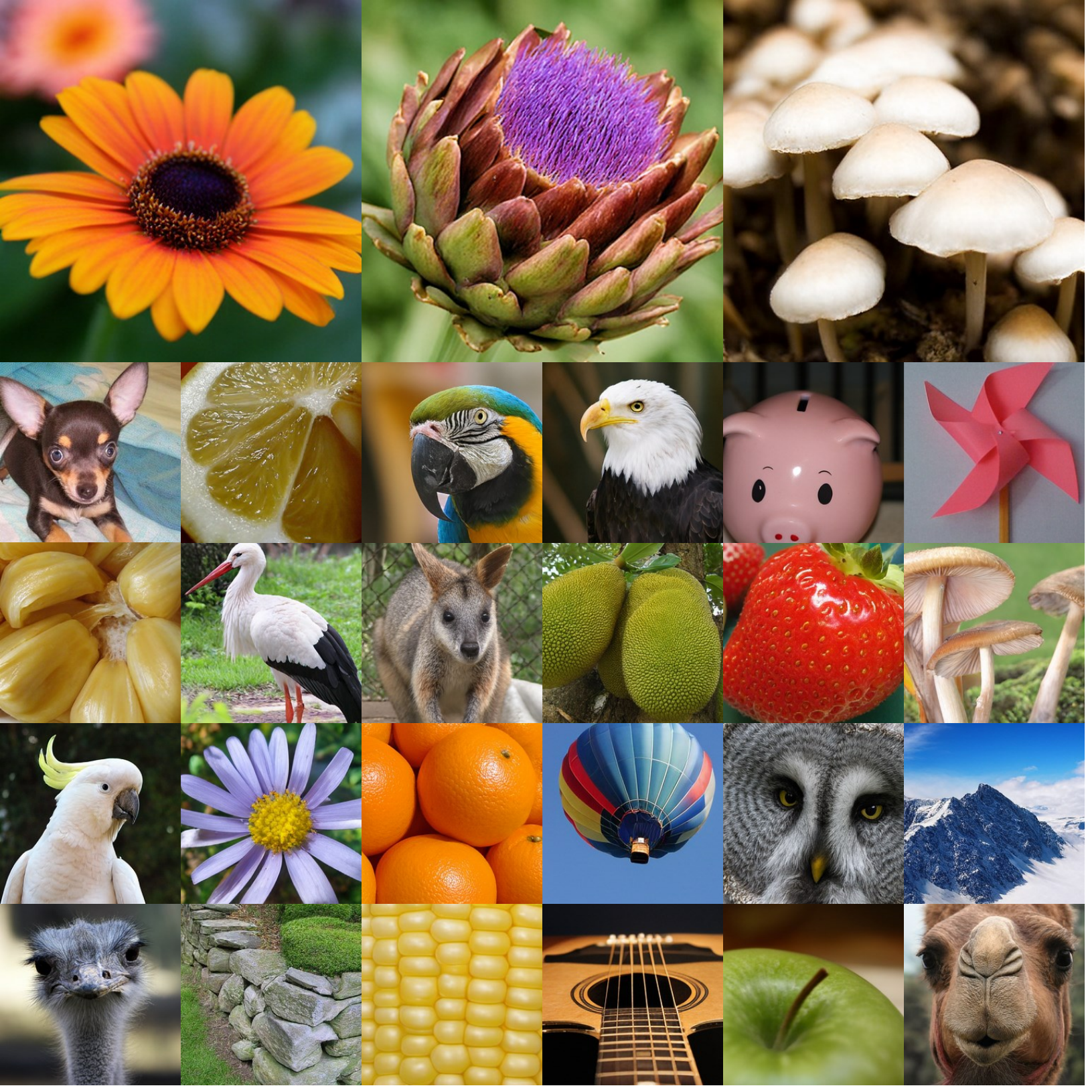}
    \caption{\textbf{Qualitative Results.} We show the images generated by our M-VAR.}
    \label{fig:qualit}
\end{figure}

We hereby train M-VAR on ImageNet~\citep{deng2009imagenet} for 512$\times$512 conditional generation. As reported in Table \ref{tab:512}, our M-VAR-d24 exhibits competitive performance in class-conditional ImageNet 512$\times$512 generation when compared to VAR. Specifically, M-VAR-d24 achieves an FID of 2.65 and an IS of 305.1, closely matching the performance of VAR-d36. More importantly, M-VAR-d24 accomplishes this with half the inference time of VAR-d36, highlighting the efficiency gains from our decoupled intra-scale and inter-scale modeling strategy. Compared to other advanced generative models, such as BigGAN, DiT-XL/2, MaskGIT, and VQGAN, M-VAR-d24 consistently outperforms them in both FID and IS metrics while maintaining a  comparable or lower inference cost. We also show more qualitative results in Figure \ref{fig:qualit}, which supports that M-VAR consistently produces images with fine details, enhanced texture fidelity, and improved structural coherence. 
\begin{table}[t]
\renewcommand\arraystretch{1.05}
\centering
\setlength{\tabcolsep}{2.5mm}{}
\small
{
\caption{
\textbf{Generative model comparison on class-conditional ImageNet 512$\times$512}.
}\label{tab:512}

\begin{tabular}{l|cc|c}
\toprule
 Model         & FID$\downarrow$ & IS$\uparrow$  &Inference Time$\downarrow$\\
\midrule
 BigGAN~\citep{biggan}  & 6.95  & 224.5    & 1     \\
 DiT-XL/2~\citep{dit} &3.04& 240.8 &160 \\
 MaskGiT~\citep{maskgit}&7.32 &156.0& 1\\
 VQGAN~\citep{vqgan}& 26.52 &66.8& 50\\
 VAR-d36~\citep{var}&2.63& 303.2&2 \\
  \rowcolor{cyan!10}
M-VAR-d24&2.65& 305.1 &1 \\
\bottomrule
\end{tabular}
}
\end{table}

\subsection{Ablation Study}
\paragraph{Parameters.} We reduce M-VAR's parameters by adjusting its width or depth, aiming for a \textit{fairer} comparison while assessing performance advantage and computational efficiency. As reported in Table \ref{tab:param}, we present three variants of M-VAR alongside the baseline VAR under similar parameter constraints. Firstly, M-VAR-W reduces the width of the model from 1024 to 768 while keeping the depth constant at 16 layers. This reduction leads to a decrease in the total number of parameters to 260 million, which is lower than VAR's 310 million parameters. Remarkably, even with fewer parameters, M-VAR-W achieves a better FID score of 3.20 compared to VAR's 3.55; Additionally, the training cost is reduced to 0.9 times that of VAR, showcasing enhanced efficiency. Similarly, M-VAR-D maintains the original width of 1024 but reduces the depth from 16 to 12 layers. M-VAR-D attains an FID score of 3.19, outperforming VAR while also reducing the training/inference cost to 0.8/0.7 times that of the VAR. These results corroborate that our proposed M-VAR models can achieve superior image generation quality compared to the baseline VAR, even when operating under similar or reduced parameter budgets.

\begin{table}[]
    \centering
    \caption{Compare with VAR under similar parameters. $\dag$ our default settings.}
    \begin{tabular}{c|c|c|c|c|c|c}
    
    \toprule
        Model & Depth & With & Param. & FID$\downarrow$ & Training Cost $\downarrow$& Inference Time$\downarrow$ \\
        \midrule
        VAR & 16& 1024&310M&3.55&1& 0.9\\
        M-VAR-W& 16& 768&260M& 3.20& 0.9& 0.9\\
        M-VAR-D& 12& 1024&340M& 3.19&0.8& 0.7\\
        M-VAR$\dag$& 16& 1024&450M& 3.07&1&1\\
        \bottomrule
    \end{tabular}
    
    \label{tab:param}
\end{table}

\begin{figure}
    \centering
    \includegraphics[width=0.5\linewidth]{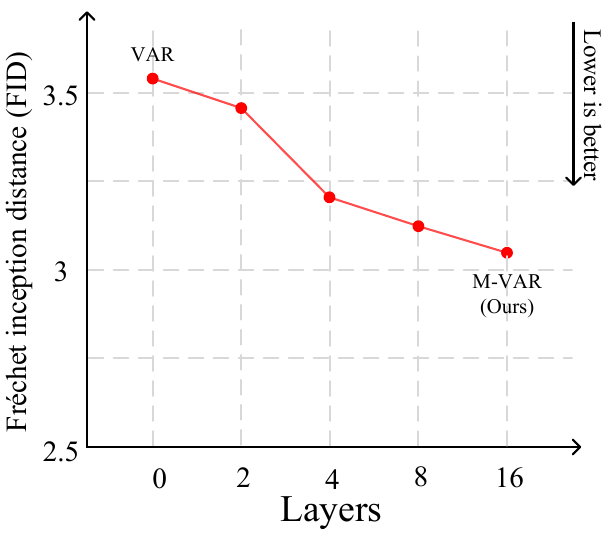}
    \caption{The effectiveness of our decouple design. We gradually replace the global attention with our intra-scale attention and inter-scale mamba.}
    \label{fig:var2mar}
\end{figure}

\paragraph{From VAR to MAR.}
We gradually replaced the global attention layers in VAR with our proposed intra-scale attention and inter-scale Mamba modules to evaluate their impact on image generation quality. As shown in Figure~\ref{fig:var2mar}, by progressively increasing the number of layers replaced, from 0 in the original VAR to all 16 layers in our M-VAR, we can observe a consistent improvement in FID scores from 3.55 to 3.07. 
This phenomenon suggests that decoupling the modeling of intra-scale and inter-scale dependencies positively impacts image synthesis quality --- by effectively capturing local spatial details within each scale and efficiently modeling hierarchical relationships between scales, our approach successfully leads to more coherent and detailed image generation.

\paragraph{Effectiveness and efficiency of Attention and Mamba.}

Table~\ref{tab:mamba} illustrates the impact of different attention mechanisms on image generation quality, as measured by FID. The baseline VAR employs global attention, capturing both intra-scale and inter-scale dependencies simultaneously, and achieves an FID of 3.55. Then, when exclusively using intra-scale attention (\ie, without inter-scale modeling, Method 1), the FID significantly deteriorates to 7.17, indicating that inter-scale dependencies are crucial for high-quality image generation. Method 2, which additionally introduces token-wise relationship modeling along all scales, improves the FID to 4.12, yet still falls short of the baseline VAR performance. Our proposed M-VAR model combines intra-scale attention with Mamba for efficient inter-scale modeling --- by decoupling the two types of dependencies and applying Mamba's linear-complexity approach for inter-scale interactions, M-VAR achieves the best FID of 3.07. This demonstrates that effectively capturing intra-scale dependencies with attention and efficiently modeling inter-scale relationships with Mamba leads to superior image quality.

\begin{table}[]
    \centering
    \caption{Effectiveness and efficiency of Attention and Mamba. We compare our intra-scale attention and Mamba with previous global attention in VAR}
    \begin{tabular}{c|c|c|c|c}
    \toprule
      Method   & Global Attention & Intra-scale Attention & Mamba & FID $\downarrow$\\
      \midrule
       VAR  &$\checkmark$&&&3.55 \\
       1&&$\checkmark$&&7.17\\
       2&&&$\checkmark$&4.12\\
        \rowcolor{cyan!10}
       M-VAR (Ours)&&$\checkmark$&$\checkmark$ &3.07 \\
    \bottomrule
    \end{tabular}
    
    \label{tab:mamba}
\end{table}

\section{Conclusion}
This paper develops a novel scale-wise autoregressive image generation method, which decouples intra-scale and inter-scale modeling to enhance both efficiency and performance. Our key idea is that, for inter-scale modeling, we replace the standard attention with Mamba for 
global-sequence modeling. This strategic separation allows our model to maintain spatial coherence and hierarchical consistency while significantly reducing computational complexity. Our experiments demonstrate that this decoupled framework outperforms existing autoregressive models and diffusion models, achieving superior image quality with fewer parameters and faster inference speeds.

\bibliography{iclr2025_conference}

\begin{thebibliography}{38}
\providecommand{\natexlab}[1]{#1}
\providecommand{\url}[1]{\texttt{#1}}
\expandafter\ifx\csname urlstyle\endcsname\relax
  \providecommand{\doi}[1]{doi: #1}\else
  \providecommand{\doi}{doi: \begingroup \urlstyle{rm}\Url}\fi

\bibitem[dit(2024)]{dit-github}
Alpha-vllm. large-dit-imagenet.
\newblock 2024.
\newblock URL \url{https://github.com/Alpha-VLLM/LLaMA2-Accessory/tree/ f7fe19834b23e38f333403b91bb0330afe19f79e/Large-DiT-ImageNet, 2024}.

\bibitem[Azadi et~al.(2018)Azadi, Olsson, Darrell, Goodfellow, and Odena]{rej2}
Samaneh Azadi, Catherine Olsson, Trevor Darrell, Ian Goodfellow, and Augustus Odena.
\newblock Discriminator rejection sampling.
\newblock \emph{arXiv preprint arXiv:1810.06758}, 2018.

\bibitem[Brock et~al.(1809)Brock, Donahue, and Simonyan]{biggan}
Andrew Brock, Jeff Donahue, and Karen Simonyan.
\newblock Large scale gan training for high fidelity natural image synthesis. arxiv 2018.
\newblock \emph{arXiv preprint arXiv:1809.11096}, 1809.

\bibitem[Brown et~al.(2020)Brown, Mann, Ryder, Subbiah, Kaplan, Dhariwal, Neelakantan, Shyam, Sastry, Askell, et~al.]{gpt3}
Tom~B. Brown, Benjamin Mann, Nick Ryder, Melanie Subbiah, Jared Kaplan, Prafulla Dhariwal, Arvind Neelakantan, Pranav Shyam, Girish Sastry, Amanda Askell, et~al.
\newblock Language models are few-shot learners.
\newblock In \emph{Advances in Neural Information Processing Systems}, volume~33, pp.\  1877--1901, 2020.
\newblock URL \url{https://proceedings.neurips.cc/paper/2020/file/1457c0d6bfcb4967418bfb8ac142f64a-Paper.pdf}.

\bibitem[Chang et~al.(2022)Chang, Zhang, Jiang, Liu, and Freeman]{maskgit}
Huiwen Chang, Han Zhang, Lu~Jiang, Ce~Liu, and William~T Freeman.
\newblock Maskgit: Masked generative image transformer.
\newblock In \emph{Proceedings of the IEEE/CVF Conference on Computer Vision and Pattern Recognition}, pp.\  11315--11325, 2022.

\bibitem[Dao(2024)]{flash2}
Tri Dao.
\newblock Flash{A}ttention-2: Faster attention with better parallelism and work partitioning.
\newblock In \emph{International Conference on Learning Representations (ICLR)}, 2024.

\bibitem[Dao \& Gu(2024)Dao and Gu]{mamba2}
Tri Dao and Albert Gu.
\newblock Transformers are ssms: Generalized models and efficient algorithms through structured state space duality.
\newblock \emph{arXiv preprint arXiv:2405.21060}, 2024.

\bibitem[Dao et~al.(2022)Dao, Fu, Ermon, Rudra, and R{\'e}]{flash1}
Tri Dao, Daniel~Y. Fu, Stefano Ermon, Atri Rudra, and Christopher R{\'e}.
\newblock Flash{A}ttention: Fast and memory-efficient exact attention with {IO}-awareness.
\newblock In \emph{Advances in Neural Information Processing Systems (NeurIPS)}, 2022.

\bibitem[Deng et~al.(2009)Deng, Dong, Socher, Li, Li, and Fei-Fei]{deng2009imagenet}
Jia Deng, Wei Dong, Richard Socher, Li-Jia Li, Kai Li, and Li~Fei-Fei.
\newblock Imagenet: A large-scale hierarchical image database.
\newblock In \emph{2009 IEEE conference on computer vision and pattern recognition}, pp.\  248--255. Ieee, 2009.

\bibitem[Devlin(2018)]{bert}
Jacob Devlin.
\newblock Bert: Pre-training of deep bidirectional transformers for language understanding.
\newblock \emph{arXiv preprint arXiv:1810.04805}, 2018.

\bibitem[Dhariwal \& Nichol(2021)Dhariwal and Nichol]{adm}
Prafulla Dhariwal and Alexander Nichol.
\newblock Diffusion models beat gans on image synthesis.
\newblock \emph{Advances in neural information processing systems}, 34:\penalty0 8780--8794, 2021.

\bibitem[Esser et~al.(2021)Esser, Rombach, and Ommer]{vqgan}
Patrick Esser, Robin Rombach, and Bjorn Ommer.
\newblock Taming transformers for high-resolution image synthesis.
\newblock In \emph{Proceedings of the IEEE/CVF conference on computer vision and pattern recognition}, pp.\  12873--12883, 2021.

\bibitem[Fei et~al.(2024)Fei, Fan, Yu, Li, Zhang, and Huang]{dim}
Zhengcong Fei, Mingyuan Fan, Changqian Yu, Debang Li, Youqiang Zhang, and Junshi Huang.
\newblock Dimba: Transformer-mamba diffusion models.
\newblock \emph{arXiv preprint arXiv:2406.01159}, 2024.

\bibitem[Grover et~al.(2018)Grover, Gummadi, Lazaro-Gredilla, Schuurmans, and Ermon]{rej1}
Aditya Grover, Ramki Gummadi, Miguel Lazaro-Gredilla, Dale Schuurmans, and Stefano Ermon.
\newblock Variational rejection sampling.
\newblock In \emph{International Conference on Artificial Intelligence and Statistics}, pp.\  823--832. PMLR, 2018.

\bibitem[Gu \& Dao(2023)Gu and Dao]{mamba1}
Albert Gu and Tri Dao.
\newblock Mamba: Linear-time sequence modeling with selective state spaces.
\newblock \emph{arXiv preprint arXiv:2312.00752}, 2023.

\bibitem[Gu et~al.(2021{\natexlab{a}})Gu, Goel, and R{\'e}]{ssm1}
Albert Gu, Karan Goel, and Christopher R{\'e}.
\newblock Efficiently modeling long sequences with structured state spaces.
\newblock \emph{arXiv preprint arXiv:2111.00396}, 2021{\natexlab{a}}.

\bibitem[Gu et~al.(2021{\natexlab{b}})Gu, Johnson, Goel, Saab, Dao, Rudra, and R{\'e}]{ssm2}
Albert Gu, Isys Johnson, Karan Goel, Khaled Saab, Tri Dao, Atri Rudra, and Christopher R{\'e}.
\newblock Combining recurrent, convolutional, and continuous-time models with linear state space layers.
\newblock \emph{Advances in neural information processing systems}, 34:\penalty0 572--585, 2021{\natexlab{b}}.

\bibitem[Ho et~al.(2022)Ho, Saharia, Chan, Fleet, Norouzi, and Salimans]{cdm}
Jonathan Ho, Chitwan Saharia, William Chan, David~J Fleet, Mohammad Norouzi, and Tim Salimans.
\newblock Cascaded diffusion models for high fidelity image generation.
\newblock \emph{Journal of Machine Learning Research}, 23\penalty0 (47):\penalty0 1--33, 2022.

\bibitem[Kang et~al.(2023)Kang, Zhu, Zhang, Park, Shechtman, Paris, and Park]{gigagan}
Minguk Kang, Jun-Yan Zhu, Richard Zhang, Jaesik Park, Eli Shechtman, Sylvain Paris, and Taesung Park.
\newblock Scaling up gans for text-to-image synthesis.
\newblock In \emph{Proceedings of the IEEE/CVF Conference on Computer Vision and Pattern Recognition}, pp.\  10124--10134, 2023.

\bibitem[Lee et~al.(2022)Lee, Kim, Kim, Cho, and Han]{rq}
Doyup Lee, Chiheon Kim, Saehoon Kim, Minsu Cho, and Wook-Shin Han.
\newblock Autoregressive image generation using residual quantization.
\newblock In \emph{Proceedings of the IEEE/CVF Conference on Computer Vision and Pattern Recognition}, pp.\  11523--11532, 2022.

\bibitem[Li et~al.(2024)Li, Yang, Wang, Qiu, Chou, Li, and Li]{aim}
Haopeng Li, Jinyue Yang, Kexin Wang, Xuerui Qiu, Yuhong Chou, Xin Li, and Guoqi Li.
\newblock Scalable autoregressive image generation with mamba.
\newblock \emph{arXiv preprint arXiv:2408.12245}, 2024.

\bibitem[Li et~al.(2023)Li, Katabi, and He]{rcg}
Tianhong Li, Dina Katabi, and Kaiming He.
\newblock Return of unconditional generation: A self-supervised representation generation method.
\newblock \emph{arXiv:2312.03701}, 2023.

\bibitem[OpenAI(2022)]{chatgpt}
OpenAI.
\newblock Introducing chatgpt.
\newblock \url{https://openai.com/blog/chatgpt/}, 2022.

\bibitem[OpenAI(2023)]{gpt4}
OpenAI.
\newblock Gpt-4 technical report.
\newblock \emph{arXiv preprint arXiv:2303.08774}, 2023.
\newblock URL \url{https://arxiv.org/abs/2303.08774}.

\bibitem[Peebles \& Xie(2023)Peebles and Xie]{dit}
William Peebles and Saining Xie.
\newblock Scalable diffusion models with transformers.
\newblock In \emph{Proceedings of the IEEE/CVF International Conference on Computer Vision}, pp.\  4195--4205, 2023.

\bibitem[Radford et~al.(2018)Radford, Narasimhan, Salimans, and Sutskever]{gpt}
Alec Radford, Karthik Narasimhan, Tim Salimans, and Ilya Sutskever.
\newblock Improving language understanding by generative pre-training.
\newblock \url{https://cdn.openai.com/research-covers/language-unsupervised/language_understanding_paper.pdf}, 2018.

\bibitem[Razavi et~al.(2019)Razavi, Van~den Oord, and Vinyals]{vqvae2}
Ali Razavi, Aaron Van~den Oord, and Oriol Vinyals.
\newblock Generating diverse high-fidelity images with vq-vae-2.
\newblock \emph{Advances in neural information processing systems}, 32, 2019.

\bibitem[Ren et~al.(2024)Ren, Li, Tu, Wang, Shu, Zhang, Mei, Yang, Wang, Wang, et~al.]{arm}
Sucheng Ren, Xianhang Li, Haoqin Tu, Feng Wang, Fangxun Shu, Lei Zhang, Jieru Mei, Linjie Yang, Peng Wang, Heng Wang, et~al.
\newblock Autoregressive pretraining with mamba in vision.
\newblock \emph{arXiv preprint arXiv:2406.07537}, 2024.

\bibitem[Rombach et~al.(2022)Rombach, Blattmann, Lorenz, Esser, and Ommer]{ldm}
Robin Rombach, Andreas Blattmann, Dominik Lorenz, Patrick Esser, and Bj{\"o}rn Ommer.
\newblock High-resolution image synthesis with latent diffusion models.
\newblock In \emph{Proceedings of the IEEE/CVF conference on computer vision and pattern recognition}, pp.\  10684--10695, 2022.

\bibitem[Sauer et~al.(2022)Sauer, Schwarz, and Geiger]{stylegan-xl}
Axel Sauer, Katja Schwarz, and Andreas Geiger.
\newblock Stylegan-xl: Scaling stylegan to large diverse datasets.
\newblock volume abs/2201.00273, 2022.
\newblock URL \url{https://arxiv.org/abs/2201.00273}.

\bibitem[Sun et~al.(2024)Sun, Jiang, Chen, Zhang, Peng, Luo, and Yuan]{llamagen}
Peize Sun, Yi~Jiang, Shoufa Chen, Shilong Zhang, Bingyue Peng, Ping Luo, and Zehuan Yuan.
\newblock Autoregressive model beats diffusion: Llama for scalable image generation.
\newblock \emph{arXiv preprint arXiv:2406.06525}, 2024.

\bibitem[Tian et~al.(2024)Tian, Jiang, Yuan, Peng, and Wang]{var}
Keyu Tian, Yi~Jiang, Zehuan Yuan, Bingyue Peng, and Liwei Wang.
\newblock Visual autoregressive modeling: Scalable image generation via next-scale prediction.
\newblock \emph{arXiv preprint arXiv:2404.02905}, 2024.

\bibitem[Touvron et~al.(2023)Touvron, Lavril, Izacard, Martinet, Lachaux, Lacroix, Rozi{\`e}re, Goyal, Hambro, Azhar, et~al.]{llama}
Hugo Touvron, Thibaut Lavril, Gautier Izacard, Xavier Martinet, Marie-Anne Lachaux, Timoth{\'e}e Lacroix, Baptiste Rozi{\`e}re, Naman Goyal, Eric Hambro, Faisal Azhar, et~al.
\newblock Llama: Open and efficient foundation language models.
\newblock \emph{arXiv preprint arXiv:2302.13971}, 2023.

\bibitem[Van~den Oord et~al.(2016)Van~den Oord, Kalchbrenner, Espeholt, Vinyals, Graves, et~al.]{pixelcnn}
Aaron Van~den Oord, Nal Kalchbrenner, Lasse Espeholt, Oriol Vinyals, Alex Graves, et~al.
\newblock Conditional image generation with pixelcnn decoders.
\newblock \emph{Advances in neural information processing systems}, 29, 2016.

\bibitem[Vaswani(2017)]{vaswani2017attention}
A~Vaswani.
\newblock Attention is all you need.
\newblock \emph{Advances in Neural Information Processing Systems}, 2017.

\bibitem[Yu et~al.(2021)Yu, Li, Koh, Zhang, Pang, Qin, Ku, Xu, Baldridge, and Wu]{vit-vqgan}
Jiahui Yu, Xin Li, Jing~Yu Koh, Han Zhang, Ruoming Pang, James Qin, Alexander Ku, Yuanzhong Xu, Jason Baldridge, and Yonghui Wu.
\newblock Vector-quantized image modeling with improved vqgan.
\newblock \emph{arXiv preprint arXiv:2110.04627}, 2021.

\bibitem[Yu et~al.(2022)Yu, Xu, Koh, Luong, Baid, Wang, Vasudevan, Ku, Yang, Ayan, et~al.]{parti}
Jiahui Yu, Yuanzhong Xu, Jing~Yu Koh, Thang Luong, Gunjan Baid, Zirui Wang, Vijay Vasudevan, Alexander Ku, Yinfei Yang, Burcu~Karagol Ayan, et~al.
\newblock Scaling autoregressive models for content-rich text-to-image generation.
\newblock \emph{arXiv preprint arXiv:2206.10789}, 2\penalty0 (3):\penalty0 5, 2022.

\bibitem[Zhu et~al.(2024)Zhu, Liao, Zhang, Wang, Liu, and Wang]{vim}
Lianghui Zhu, Bencheng Liao, Qian Zhang, Xinlong Wang, Wenyu Liu, and Xinggang Wang.
\newblock Vision mamba: Efficient visual representation learning with bidirectional state space model.
\newblock \emph{arXiv preprint arXiv:2401.09417}, 2024.

\end{thebibliography}
\bibliographystyle{iclr2025_conference}

\end{document}